\documentclass[conference]{IEEEtran}
\IEEEoverridecommandlockouts
\usepackage{cite}
\usepackage{amsmath,amssymb,amsfonts}
\usepackage{algorithm}
\usepackage{algorithmic}
\usepackage{graphicx}
\usepackage{textcomp}
\usepackage{multirow}
\usepackage{xcolor}
\usepackage{tabularx}
\usepackage{float}
\usepackage{threeparttable}
\usepackage[colorlinks=true, urlcolor=blue]{hyperref}
\def\BibTeX{{\rm B\kern-.05em{\sc i\kern-.025em b}\kern-.08em
    T\kern-.1667em\lower.7ex\hbox{E}\kern-.125emX}}
\begin{document}

\title{FocusMed: A Large Language Model-based
Framework for Enhancing Medical Question
Summarization with Focus Identification
}

\author{
\IEEEauthorblockN{
Chao Liu\textsuperscript{1},
Ling Luo\textsuperscript{1, *},
Tengxiao Lv\textsuperscript{1},
Huan Zhuang\textsuperscript{2},
Lejing Yu\textsuperscript{2},
Jian Wang\textsuperscript{1},
Hongfei Lin\textsuperscript{1}
}
\IEEEauthorblockA{\textsuperscript{1} College of Computer Science and Technology,
Dalian University of Technology, Dalian, China}
\IEEEauthorblockA{\textsuperscript{2} Cancer Hospital of Dalian University of Technology, Liaoning Cancer Hospital \& Institute, Shenyang, China}
\IEEEauthorblockA{*To whom correspondence should be addressed: lingluo@dlut.edu.cn}
}

\maketitle

\begin{abstract}
With the rapid development of online medical platforms, consumer health questions (CHQs) are inefficient in diagnosis due to redundant information and frequent non-professional terms. The medical question summary (MQS) task aims to transform CHQs into streamlined doctors’ frequently asked questions (FAQs), but existing methods still face challenges such as poor identification of question focus and model hallucination. This paper explores the potential of large language models (LLMs) in the MQS task and finds that direct fine-tuning is prone to focus identification bias and generates unfaithful content. To this end, we propose an optimization framework based on core focus guidance. First, a prompt template is designed to drive the LLMs to extract the core focus from the CHQs that is faithful to the original text. Then, a fine-tuning dataset is constructed in combination with the original CHQ-FAQ pairs to improve the ability to identify the focus of the question. Finally, a multi-dimensional quality evaluation and selection mechanism is proposed to comprehensively improve the quality of the summary from multiple dimensions. We conduct comprehensive experiments on two widely-adopted MQS datasets using three established evaluation metrics. The proposed framework achieves state-of-the-art performance across all measures, demonstrating a significant boost in the model’s ability to identify critical focus of questions and a notable mitigation of hallucinations. The source codes are freely available at \href{https://github.com/DUT-LiuChao/FocusMed}{https://github.com/DUT-LiuChao/FocusMed}.
\end{abstract}


\section{Introduction}
With the advancement of science and technology and the continuous development of artificial intelligence, online medical platforms are becoming increasingly popular among consumers. These platforms allow users to consult various health-related issues anytime and anywhere. However, consumer questions are typically expressed in natural language. They often contain excessive, irrelevant information and unprofessional vocabulary, which complicates diagnosis. These factors significantly increase the difficulty of identifying user needs and generating accurate responses. To address this issue, Abacha \cite{abacha2019summarization} proposed the task of Medical Question Summary (MQS). The core goal of MQS is to summarize the lengthy consumer health questions (CHQs) into concise, general questions suitable for doctors. These summaries are typically short, enabling doctors to understand user intent more efficiently and provide more accurate and relevant answers.

Since Abacha \cite{abacha2019summarization} introduced the MQS task in 2019, research in this area has continued to expand, leading to the development of various deep learning-based methods. These methods can be broadly divided into three categories: sequence-to-sequence (Seq2Seq)-based methods, reinforcement learning-based methods, and contrastive learning-based methods. Early Seq2Seq-based methods \cite{sutskever2014sequence} mainly used the encoder-decoder architecture combined with the attention mechanism to handle medical quesitons and generate summaries. However, due to the complexity and professionalism of medical question, traditional Seq2Seq methods often struggle with domain-specific knowledge and are prone to generating inaccurate or irrelevant summaries \cite{liu2019text}. With the emergence of pre-trained language models (PLMs), variant Seq2Seq models have become widely used in MQS tasks. Models such as BART \cite{lewis2019bart} have shown strong performance, helping to mitigate the limitations of earlier Seq2Seq methods in handling specific biomedical knowledge. Nevertheless, most existing PLMs use maximum likelihood estimation (MLE), which focuses on the accuracy of masked tag predictions but does not fully capture the semantic similarity or dissimilarity of the entire sentence \cite{zhang2022focus}. To address this quesiton, some studies have introduced reinforcement learning \cite{yadav2021reinforcement} in text summarization tasks. Experiments have shown that applying contrastive learning with positive and negative samples \cite{wei2023medical}, can enhance the model's semantic representations and improve its performance on MQS tasks. 

While these methods have achieved promising results in this task, they are still limited by the pre-training corpus of PLMs. This limitation makes it difficult to effectively capture complex semantic information. In recent years,  advancements in large language models (LLMs), such as GPT-4 \cite{achiam2023gpt}, Med-PaLM 2 \cite{singhal2025toward} and Taiyi \cite{luo2024taiyi}, have attracted significant attention in the medical field, offering new opportunities for improving the performance of MQS tasks. Compared with previous models, LLMs demonstrate significantly better performance. Their advantages stem from the use of large-scale corpora and powerful computational resources \cite{yu2024enhancing}, which enable them to better understand and process complex language structures. As a result, LLMs can effectively capture intricate semantic information in medical question-and-answer tasks, allowing them to generate more accurate and contextually appropriate summaries \cite{diekmann2025evaluating}. In addition, LLMs acquire rich domain knowledge during pre-training, which can be effectively transferred to a wide range of downstream tasks, especially in the medical domain \cite{nori2023capabilities}. With only a small amount of fine-tuning data, these models can quickly adapt to specific medical applications, significantly improving the quality and accuracy of their outputs.

\begin{figure}[t]
\centering
\includegraphics[width=0.5\textwidth]{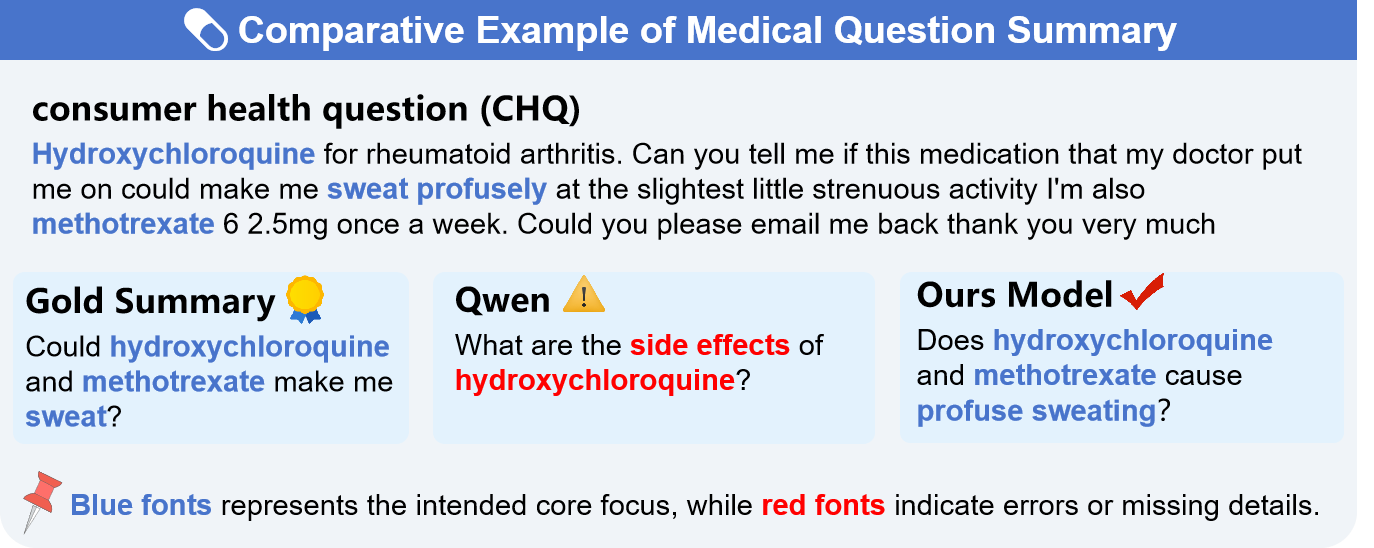}
\caption{A Comparative example for medical question summarization in MEDIQA dataset. Qwen represents the output generated directly by the fine-tuned Qwen2.5-7B model, while Our Model refers to the result obtained through the proposed FocusMed.}
\label{fig:description}
\end{figure}

Although LLMs have shown promising prospects in the medical field, their inability to accurately identify or maintain focus in MQS tasks remains a major limitation \cite{schumacher2023extrinsically}. As shown in Fig. \ref{fig:description}, even after fine-tuning Qwen2.5-7B \cite{yang2025qwen3} on a domain-specific dataset, the model still exhibits certain shortcomings. Compared to the gold-standard summary, the results generated by Qwen not only omit the important medication ``methotrexate'', but also misinterpret the patient's inquiry about the correlation between the medication and symptoms as a question about drug side effects.

In addition, hallucination remains a prominent issue when LLMs are used to generate medical question summaries. Hallucination refers to the generation of irrelevant or inaccurate information by the model during text generation, which is particularly critical in medical tasks where inaccurately generated content may lead to serious consequences \cite{pham2024towards}. For example, an LLM might incorrectly generate non-existent medications, side effects, or treatments \cite{pal2023med}. Such errors undermine the model's practical applicability and reliability. Therefore, mitigating hallucinations and improving the accuracy and relevance of the model's generated content remain critical challenges in the medical application of LLMs \cite{zuo2024medhallbench}.

To address these issues, we propose FocusMed, an LLM-based framework designed to enhance MQS tasks by incorporating focus identification. First, we use LLMs to extract the core focus of each consumer health question. Specifically, we design precise prompts to generate core focus points from the CHQs, combined with a faithfulness validation mechanism to reduce hallucinations in LLMs. Next, we integrate the extracted core focuses with the original CHQs to construct a new enhanced dataset with clearer structure and more explicit semantics. This augmented dataset serves as a foundation for training subsequent models, aiming to improve their ability to identify key focus points effectively. Building on this, we propose a multi-dimensional quality evaluation and selection mechanism. By combining the complementary strengths of different models, this selection mechanism boosts the overall performance and improves the accuracy and quality of the generated summaries. It also reduces the impact of individual model errors, ensuring more consistent and reliable results.

Our main contributions are summarized as follows:
\begin{itemize}
\item[$\bullet$]   We propose the FocusMed framework. It begins by using LLMs to extract the core focus of CHQs and construct an enhanced dataset to improve the model’s focus identification capabilities. Then, multi-dimensional quality evaluation and selection mechanism is introduced to further enhance the overall system's performance.
\item We evaluate the performance of both open-source and proprietary LLMs on MQS tasks. By employing various strategies, such as few-shot prompting and parameter-efficient fine-tuning techniques like QLoRA \cite{dettmers2023qlora}, we conduct a comprehensive assessment of state-of-the-art LLMs. 
\item[$\bullet$]  We validate the effectiveness of our framework on two benchmark datasets. Our FocusMed model achieves ROUGE-L improvements of 7.3\% and 6.7\% over the current state-of-the-art methods on the MEDIQA and MeqSum datasets, respectively. Additionally, the generated summaries demonstrate improved faithfulness.
\end{itemize}
\section{Methods}
\begin{figure*}[htbp]
    \centering
    \includegraphics[width=1\textwidth]{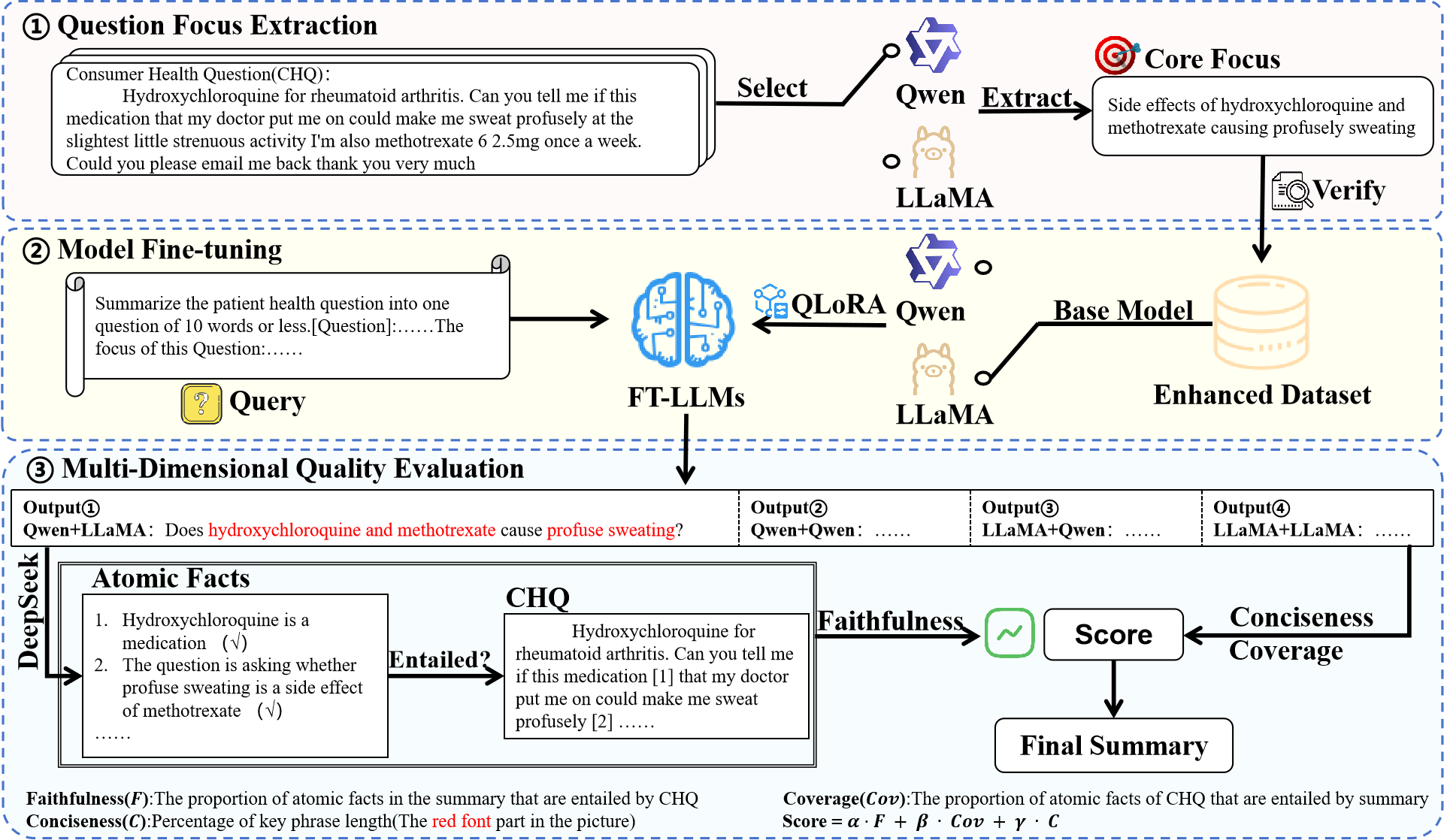}
    \caption{The overall framework of our FocusMed. We first utilize LLMs to extract key focuses from CHQ questions, constructing an enhanced dataset for subsequent fine-tuning. During the extraction and fine-tuning stages, we generate results using four different model combinations based on Qwen2.5-7B and LLaMA3.1-8B. (e.g., ``Qwen+LLaMA'' indicates that Qwen is used in the extraction stage and LLaMA in the fine-tuning stage.) The final outputs are selected based on three dimensions: faithfulness, conciseness, and coverage. Since the calculation process of coverage and faithfulness is similar, it is not shown in the figure.}
    \label{fig:process}
\end{figure*}

First, we briefly introduce the definition of the MQS task  (§\ref{sec:description}). Then, we detail the FocusMed framework proposed to address the challenge of poor focus identification, as illustrated in Fig. \ref{fig:process}. The framework consists of three main components: Question Focus Extraction (§\ref{sec:main}), Model Fine-tuning (§\ref{sec:Fine-tuning}), and a Multi-Dimensional Quality Evaluation and Selection Mechanism (§\ref{sec:ensemble}).
\subsection{Task Definition}\label{sec:description}
The primary objective of the MQS task is to convert lengthy and complex consumer health queries into concise, standardized questions that are easier for doctors to comprehend. By extracting critical information from detailed and often convoluted inputs, this process enables healthcare professionals to efficiently identify the user’s core concerns and provide more accurate and timely medical advice.
\subsection{Question Focus Extraction}\label{sec:main}
\begin{figure}[t]
    \centering
    \includegraphics[width=0.42\textwidth]{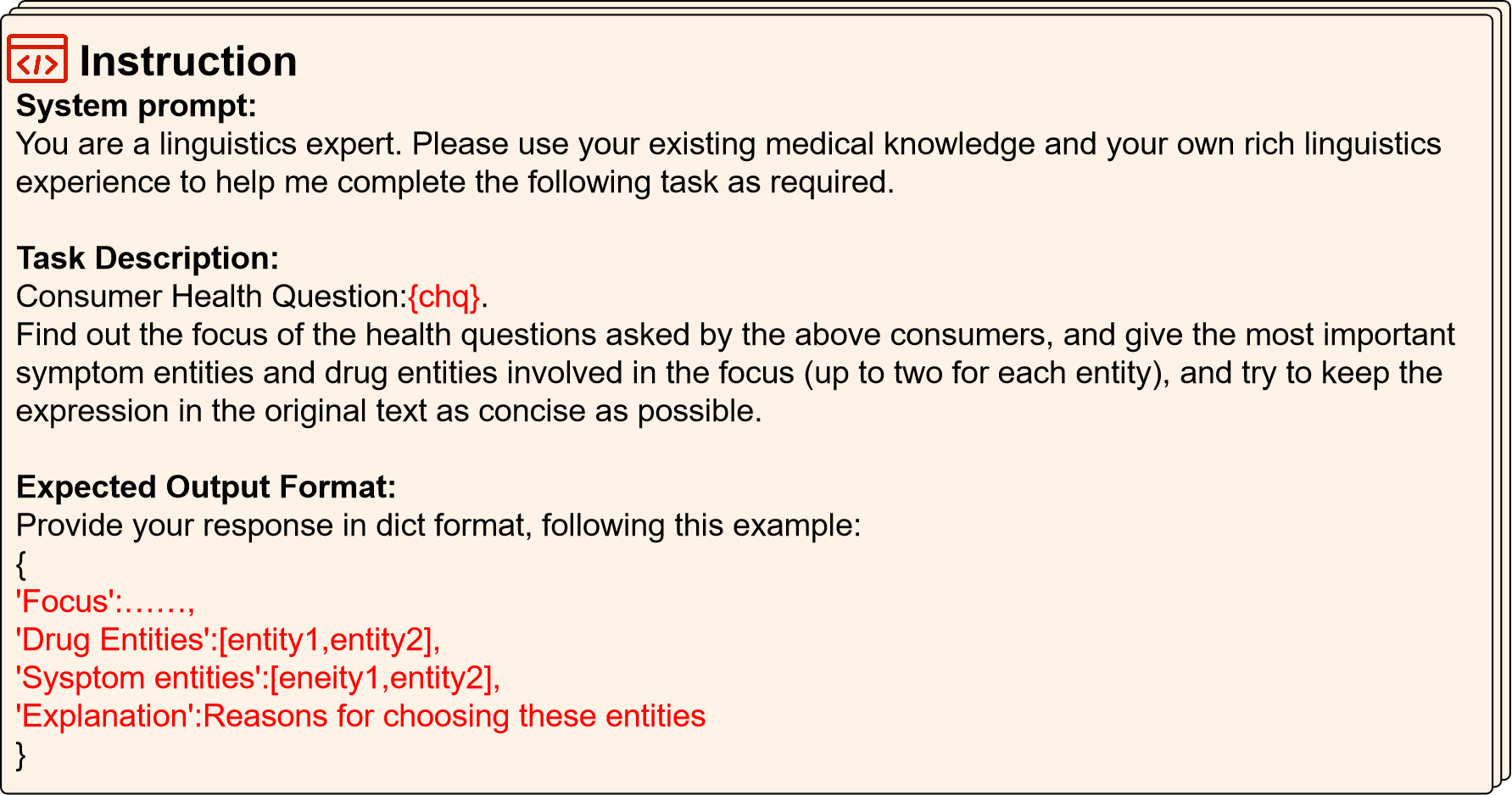}
    \caption{Instructions used in the question focus extraction phase.}
    \label{fig:instruction}
\end{figure}

Although supervised fine-tuning methods can enhance model performance in MQS tasks \cite{savage2024fine}, the persistent challenge of accurately identifying the question focus remains unresolved. To address this, we propose a lightweight optimization approach based on prompt engineering that leverages the inherent capabilities of LLMs. Without relying on external resources, it deeply mines the semantic potential of the model pre-training corpus to build a self-consistent question focus extraction mechanism.
First, we design specific instructions to stimulate the internal knowledge of LLMs. (The detailed instruction design is shown in Fig. \ref{fig:instruction}). Since most CHQs focus on medications and symptoms, we instruct the model to extract only these two types of entities, with a maximum limit of two entities per category. This constraint is determined through extensive experimentation and is found to yield optimal results. Furthermore, the work of Xie \cite{xie2023doclens} demonstrated that models perform better when prompted to generate output in a structured format. Based on this finding, we adopt the JSON+CoT format in our instruction design. The model is required to output responses in JSON format and to provide justifications for its generated content. This approach helps improve the model’s reasoning ability and overall accuracy of the extracted information.

This instruction aims to extract the core requirements and relevant focus points from lengthy health-related questions. To ensure the faithfulness of the generated content and effectively mitigate hallucination issues, we introduce a semantic similarity-based validation method. Specifically, we first apply the Textrank algorithm  \cite{mihalcea2004textrank} (TextRank is a graph-based algorithm for keyword extraction. It models words as nodes, builds a graph based on co-occurrence, and uses a PageRank-like iterative process to score node importance, selecting top-ranking words as key phrases.) to extract key phrases from the model's output. Then, we traverse these key phrases and identify the most similar noun phrases from the original text, calculating their similarity. If the similarity of a key phrase is below a predefined threshold (experiments on the development set indicate that a threshold between 0.8 and 0.9 is optimal), We consider the model's output unfaithful to the original text. Conversely, if the similarity is above the threshold, we deem it faithful to the original. Finally, we integrate the generated question focus descriptions with the original training data to build an enhanced dataset, thereby improving the model's focus recognition capability. Algorithm \ref{The construct of dataset} presents the detailed workflow for this section.

\subsection{Model Fine-tuning}\label{sec:Fine-tuning}
To enable the model to more accurately extract the key focus of the question and effectively mitigate hallucination phenomena, i.e., the generation of irrelevant or false information, we select Qwen2.5-7B and LLaMA3.1-8B \cite{dubey2024llama} as the\begin{algorithm}
\caption{Construction of the enhanced dataset}
\label{The construct of dataset}
\begin{algorithmic}[1]
\REQUIRE A pretrained LLM $M$, a set of language inputs $D_{\text{input}} = \{x_i,y_i\}$, a dataset $D_i^F$, and a pre-set threshold $\tau$
\STATE \textcolor{gray}{\# Initialize dataset $D_i^F$}
\FOR{each $x_i \in D_{\text{input}}$}
    \STATE \textcolor{gray} {\# Extract the focus}
    \STATE $f_i \gets \text{Use model } M \text{ to extract the focus of } x_i$
    \STATE \textcolor{gray}{\# extract key phrase by TextRank}
    \STATE $K_F \gets \text{TextRank}(f_i)$ 
    \STATE \textcolor{gray}{\#Extract noun phrases $K_N$ from the original text}
    \FOR{each $k_f \in K_F$}
        \FOR{each $k_n \in K_N$}
            \STATE $s \gets \text{Sim}(k_n, k_f)$
            \IF{$s < \tau$}
                \STATE return to step3
            \ENDIF
        \ENDFOR
    \ENDFOR
    \STATE \textcolor{gray} {\# Combine the original text with the question focus}
    \STATE $e_i \gets x_i \bigcup f_i $
    \STATE \textcolor{gray} {\# Creating an augmented dataset}
    \STATE $D_i^F \gets D_i^F \bigcup \{e_i,y_i\} $
\ENDFOR
\end{algorithmic}
\end{algorithm} base models. The QLoRA fine-tuning technique is applied to optimize these models. This approach combines quantization with low-rank adapter techniques to reduce computational overhead and storage requirements during fine-tuning. At the same time, it preserves the expressive capacity of the models.

The dataset used is the enhanced dataset generated in §\ref{sec:main}. Specifically, each training sample consists of a pair \( (x, y) \), where the input \( x = (I, C, K) \) includes the specific task instruction \( I \), the corresponding consumer health question \( C \), and the extracted key focus \( K \), while the output \( y \) is the target summary from the dataset.

The generation process can be modeled as a series of stepwise token prediction actions. Specifically, the output sequence is denoted as \( y = [a_1, a_2, ..., a_L] \), where each token \( a_t \) is sampled from the policy \( \pi_\theta(· | s_t) \), the state \( s_t \) is composed of the input text and the tokens generated so far, and \( L \) represents the total count of tokens within the generated sequence \( y \).

\begin{equation}s_t=\left\{
\begin{array}
{ll}x, & t=0 \\
\left[x,a_1,\ldots,a_t\right], & 1\leq t\leq L
\end{array}\right.\end{equation}

The generation process terminates when the model outputs the special end-of-sequence token. During training, the objective is to minimize the negative log-likelihood loss over the silver-standard dataset \( \mathcal{D}_{\text{SFT}} \). This training objective optimizes the model's prediction probabilities on the silver-standard dataset, guiding the model to generate content that aligns with the target summaries, thereby improving its performance on the specific task.
\begin{equation}\mathcal{L}_{\mathrm{SFT}}(\theta)=-\mathrm{E}_{(x,y)\sim\mathcal{D}_{\mathrm{SFT}}}\left[\sum_{t=1}^L\log\pi_\theta(a_t\mid s_t)\right]\end{equation}
\subsection{Multi-Dimensional Quality Evaluation and Selection}\label{sec:ensemble}
To further enhance the accuracy and diversity of generated summaries, we propose a multi-dimensional evaluation and selection mechanism, as shown in Fig. \ref{fig:process}. Specifically, for a given input, we generate multiple outputs using different combinations of models for focus extraction and fine-tuning (In this experiment, we use four combinations). Each output is then evaluated and scored based on three criteria: faithfulness, conciseness, and content coverage. The output with the highest weighted score is selected as the final output of the system. The following are the detailed calculation methods of these three evaluation indicators:
\begin{itemize}
    \item [$\bullet$] Faithfulness: To evaluate the factual consistency between the model-generated summary and the original source, we adopt a fine-grained assessment approach. Specifically, we employ DeepSeek-R1  \cite{guo2025deepseek} to decompose the outputs generated by different model combinations into atomic facts, which are the most basic, indivisible factual units. DeepSeek-R1 is further used to determine which of these atomic facts are entailed by the original input. Atomic facts that are entailed are considered faithful to the source text, while those that are not are deemed unfaithful. Finally, the proportion of entailed atomic facts is taken as the faithfulness score.
    \begin{equation}
    F = \frac{N_{{match}}}{N_{{total}}}
    \end{equation}
    among them, $N_{{match}}$ represents the number of atomic facts covered in the original text, and $N_{{total}}$ represents the total number of atomic facts extracted from the summary. 
    \item [$\bullet$] Conciseness: Conciseness is an important criterion for evaluating the quality of a summary. However, it does not imply that the shorter text is always better. Rather, a concise summary should cover all essential information using as few words as possible. For this purpose, we designed the following method to measure the degree of conciseness between different results. For a given set of results obtained from different model combinations, we first apply the TextRank algorithm to extract the key phrases. Next, we calculate the sum of the lengths of these key phrases and divide it by the total length of the sentence. This ratio serves as the conciseness score for the entire sentence, providing a quantitative measure of how efficiently the summary conveys the essential information.
    \begin{equation}
     C = \frac{\sum_{i=1}^{n} L_{k_i}}{L_{total}}
    \end{equation}
    where $L_{total}$ represents the total length of the model output summary, and $L_{k_i}$ represents the length of the \textit{i}-th key phrase. $n$ is the total number of key phrases.
    \item [$\bullet$] Coverage: This metric is introduced to evaluate the extent to which the model-generated summary covers the information present in the original text. This score reflects how well the summary encapsulates the essential information from the source. Its calculation method is similar to that of the faithfulness metric. The key difference lies in the computational process: we first decompose CHQ into atomic facts, then determine which atomic facts are covered by the summary generated by the model. Let \( N_{{match}} \) be the number of atomic facts entailed by the summary, and \( N_{{total}} \) be the total number of atomic facts in the original text. The coverage score \( Cov \) is defined as the ratio of \( N_{{match}} \) to \( N_{{total}} \):
    \begin{equation}
    Cov = \frac{N_{{match}}}{N_{{total}}}
    \end{equation}
\end{itemize}

Finally, we perform a weighted calculation on the scores of the above three dimensions and select the output with the highest overall score as the final result.
\begin{equation}
    \text{Score} = \alpha \cdot F + \beta \cdot C + \gamma \cdot \mathrm{Cov}
\end{equation}

Where, $F$, $C$, $Cov$ represent the faithfulness, conciseness, and coverage scores, respectively, while $\alpha$, $\beta$, and $\gamma$ are their corresponding weights, with the constrain that $\alpha$+$\beta$+$\gamma$=1. We determined the optimal weight configurations for the MEDIQA and MeqSum datasets through grid search. For the MEDIQA dataset, the values are 0.6, 0.1, and 0.3. For the MeqSum dataset, the  values are 0.3, 0.4, and 0.3.
\section{EXPERIMENTS}
\subsection{Experiment Setup}
\subsubsection{Datasets}
To evaluate the effectiveness of our proposed method for the MQS task, we conducted experiments on two widely used MQS datasets. The first is the MeqSum dataset \cite{abacha2019summarization}, originally created by the initiators of the MQS task. It contains a total of 1,000 patient health questions, each of which is annotated with a corresponding summary by a medical expert. The second is the MEDIQA dataset \cite{abacha2021overview}, specifically the health question summary task from the MEDIQA 2021 shared task 1. Its training set comes from the MeqSum dataset, while the validation set and test set include various CHQs and their expert-generated golden summaries. Some statistics of these datasets are shown in Table \ref{Datasets}. 
\begin{table}[t]
\caption{Statistics of two medical question summarization datasets.}
\centering
{\fontsize{4pt}{5pt}\selectfont 
\resizebox{\linewidth}{!}{%
\begin{tabular}{l|c|c|c|c}
\hline
\textbf{Datasets} & \textbf{Train} & \textbf{Dev} & \textbf{Test} & \textbf{Length} \\
\hline
MEDIQA & 1000 & 50 & 100 & 66.2/11.3 \\
MeqSum & 400 & 100 & 500 & 59.4/10.0 \\
\hline
\end{tabular}
}}
\begin{tablenotes}
    \item \textbf{Notes}:  Length indicates the average length of CHQs/FAQs.
\end{tablenotes}
\label{Datasets}
\end{table}

\subsubsection{Hyperparameters}The model is trained for 10 epochs with a learning rate of 1e-4, a training batch size of 4, and a weight decay of 0.01. The maximum output length is set to 512. We conduct all experiments on a single NVIDIA 4090 24GB GPU. We use the AdamW optimizer and apply gradient clipping with a threshold of 0.3 to ensure training stability. Unless otherwise specified, all hyperparameters remain consistent across experiments.

\subsubsection{Metrics}
To evaluate the system’s output, we adopt a dual-perspective approach: General Quality and General Faithfulness. General Quality is measured using traditional metrics such as ROUGE \cite{lin2004rouge} and BERTScore \cite{zhang2019bertscore}, which assess the lexical and semantic similarity between the generated output and the reference text. General Faithfulness is evaluated using the $\textsc{SummaC}_\text{ZS}$ \cite{laban2022summac} metric, which is specifically designed for the automatic assessment of summarization quality. $\textsc{SummaC}_\text{ZS}$  focuses on the consistency and logical coherence of the generated content, ensuring that the summary accurately preserves key information from the source while maintaining structural integrity. This two-tiered evaluation framework provides a comprehensive and robust assessment of system performance.
\subsection{Experiment Results and Analysis}
\subsubsection{Performance of Different LLMs on the MQS Task}
To evaluate the performance of LLMs on the MQS task, we selected several state-of-the-art models, including open-source models such as Qwen, LLaMA, and DeepSeek-R1-Distill-Qwen, as well as the proprietary GPT-4o. To enhance model performance in these scenarios, we adopted a range of techniques, including Zero-shot prompting, chain-of-thought (CoT) reasoning, and QLoRA-based fine-tuning. These methods were designed to guide the models in generating clinically relevant and factually accurate medical summaries.

As shown in Table \ref{Explore}, on the HQS sub-task of the MEDIQA benchmark, the proprietary GPT-4o model with hundreds of billions of parameters does not demonstrate a statistically significant advantage over the open-source LLaMA3.1 model with 8 billion parameters. These findings suggest that for domain-specific tasks, model adaptability and domain-specific fine-tuning may be more effective than simply increasing model size, particularly in terms of computational efficiency and prediction accuracy. \begin{table}[b]
\caption{Performance of the LLMs on the MEDIQA dataset.}
\centering
\resizebox{\linewidth}{!}{%
\setlength{\arrayrulewidth}{0.8pt}  
\renewcommand{\arraystretch}{1.2}   
\begin{tabular}{lc|c|c|c}
\hline
\textbf{Model} & \textbf{Setting} & \textbf{ROUGE-1} & \textbf{ROUGE-2} & \textbf{ROUGE-L} \\
\hline
\multirow{2}{*}{GPT-4o}  & Zero-shot & 0.358 & 0.130 & 0.309 \\
& CoT & 0.363 & 0.135 & 0.316 \\
\hline
\multirow{3}{*}{Qwen2.5-7B} & Zero-shot      & 0.357 & 0.133 & 0.306    \\   
                            & CoT        & 0.346 & 0.135 & 0.303 \\
                            & SFT      & 0.366 & 0.158 & 0.334 \\
\hline
\multirow{3}{*}{DS-Qwen-7B} & Zero-shot      & 0.311 & 0.104 & 0.253    \\
                            & CoT & 0.327 & 0.119 & 0.280 \\
                            & SFT      & 0.330 & 0.131 & 0.299 \\       
\hline
\multirow{3}{*}{LLaMA3.1-8B}& Zero-shot      & 0.364 & 0.130 & 0.312    \\ 
                            & CoT & 0.372 & 0.133 & 0.326 \\
                            & SFT      & \textbf{0.379} & \textbf{0.160} & \textbf{0.339} \\
\hline
\end{tabular}
}
\begin{tablenotes}
\item \textbf{Notes}: DS-Qwen-7B refers to DeepSeek-R1-Distill-Qwen-7B.
\end{tablenotes}
\label{Explore}
\end{table}Moreover, within the same model architecture, fine-tuning consistently outperforms zero-shot and CoT, providing empirical evidence to support this conclusion.

Although DeepSeek-R1-Distill-Qwen-7B exhibits strong performance on tasks requiring advanced reasoning, such as mathematics, its performance on the MQS task is notably inferior to that of other models of comparable size. This discrepancy is primarily attributed to the model being distilled from Qwen2.5-Math-7B, which has been fine-tuned with additional mathematical data. While such task-specific fine-tuning enhances performance in mathematical reasoning, it may compromise the model’s generalization capabilities on broader natural language tasks, including sentiment analysis, summarization, and commonsense reasoning. This observation further underscores the importance of model adaptability, as discussed previously.

\subsubsection{Effect of Question Focus Extraction}
In this section, we validate the effectiveness of the proposed method described in §\ref{sec:main}. Two models with comparable parameter sizes, Qwen2.5-7B and LLaMA3.1-8B, were selected for experimentation. During the question focus extraction and fine-tuning stages, one of the models was randomly chosen to construct different model combinations. The experimental results are summarized in Table \ref{tab:two stage results}. The results clearly indicate that incorporating question focus information extracted in the first stage significantly improves model performance. This improvement is particularly pronounced when LLaMA3.1-8B serves as the fine-tuning base. Furthermore, the introduction of faithfulness verification substantially enhances the $\textsc{SummaC}_\text{ZS}$ score.

\newcolumntype{C}[1]{>{\centering\arraybackslash}p{#1}}
\begin{table}[t]
\caption{Experimental results of question focus extraction method.}
\centering
{\fontsize{14pt}{18pt}\selectfont
\setlength{\arrayrulewidth}{0.5mm}
\resizebox{\linewidth}{!}{%
\begin{tabular}{l|l|c|c|c}
\hline
 \textbf{Extraction Stage} &\textbf{Fine-tune Stage} & \textbf{ROUGE-L} & \textbf{BERTScore} & \textbf{$\textsc{SummaC}_\text{ZS}$} \\
\hline
—— & Qwen2.5-7B   & 0.334 & 0.765 & 0.518 \\
—— & LLaMA3.1-8B  & 0.339 & 0.767 & 0.526 \\
\hline
Qwen2.5-7B & \multirow{2}{*}{Qwen2.5-7B}   & 0.347 & 0.765 & 0.563 \\
LLaMA3.1-8B     & & 0.345 & 0.766 & 0.541 \\
\hline
Qwen2.5-7B  & \multirow{2}{*}
{LLaMA3.1-8B}   & \textbf{0.364} & 0.775 & \textbf{0.577} \\
LLaMA3.1-8B & & 0.362 & \textbf{0.778} & 0.572 \\
\hline
\end{tabular}%
}}
\label{tab:two stage results}
\end{table}

\begin{figure}[b]
\centering
\includegraphics[width=0.5\textwidth]{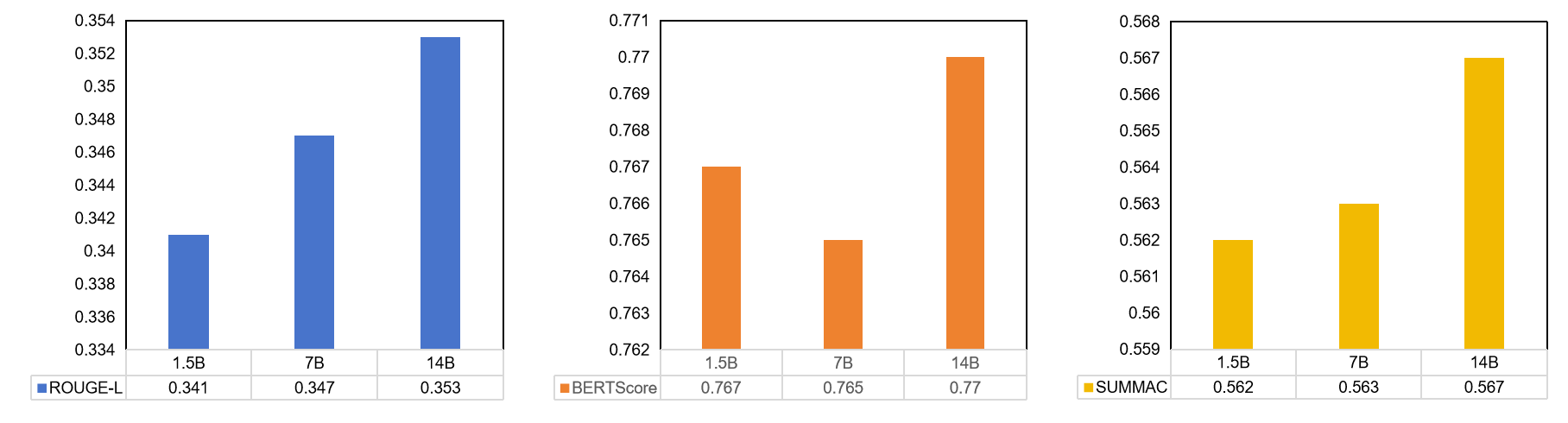}
\caption{Performance of LLMs with different sizes in the extraction stage.}
\label{fig:parameter sizes}
\end{figure}
We further investigated the impact of model parameter size during the question focus identification stage. Specifically, in the extraction phase, we employed Qwen2.5 models with 1.5B, 7B, and 14B parameters, while using Qwen2.5-7B as the fine-tuning base. As shown in Fig. \ref{fig:parameter sizes}, increasing the number of parameters in the extraction model leads to improved performance. This improvement is attributed to the larger models’ access to broader pre-training corpora, enabling them to acquire richer medical knowledge and stronger information extraction capabilities. Consequently, these models more accurately identify the question focus, addressing the limitations of smaller models in capturing key information.

\subsubsection{Effect of Multi-Dimensional Quality Evaluation}
In this section, we evaluate the effectiveness of the multi-dimensional quality evaluation and selection mechanism proposed in §\ref{sec:ensemble}. Qwen2.5-7B and LLaMA3.1-8B are selected, and by varying the model combinations used during the extraction and fine-tuning stages, four distinct outputs are generated. These outputs are then evaluated based on three criteria: faithfulness, conciseness, and content coverage. The output with the highest overall score is selected as the final system output. The experimental results are presented in Table \ref{tab:result of ensemble}, which also includes comparisons excluding the question focus extraction stage. This demonstrates that our framework is not merely a simple aggregation of two components, but rather a tightly integrated system with synergistic effects. The results clearly show that the proposed selection mechanism significantly enhances overall system performance.
\newcolumntype{C}[1]{>{\centering\arraybackslash}p{#1}}
\begin{table}[b]
  \centering
  \caption{Experimental results of Multi-Dimensional Quality Evaluation and Selection on MEDIQA dataset.}
  {\fontsize{6pt}{8pt}\selectfont
  \setlength{\arrayrulewidth}{0.23mm}
  \resizebox{\linewidth}{!}{%
  \begin{tabular}{l|c|c|c}
    \hline
\textbf{Model} & \textbf{ROUGE-L} & \textbf{BERTscore} & \textbf{$\textsc{SummaC}_\text{ZS}$}  \\
    \hline
    LLMs Selection &0.371&0.775&0.596\\
    FocusMed(w/o Selection) & 0.364 & 0.775 & 0.577 \\
    FocusMed(w/o Extract)   & 0.347   & 0.768 & 0.548  \\
    FocusMed    & \textbf{0.386} & \textbf{0.782} & \textbf{0.603}\\
    \hline
  \end{tabular}
  }}
\begin{tablenotes}
\item \textbf{Notes}: ``LLMs Selection'' means that we directly use LLM to select the summary it deems the best. ``FocusMed(w/o Selection)'' refers to the best result among the four models used for the selection. ``FocusMed(w/o Extract)'' refers to the use of directly fine-tuned Qwen and LLaMA to obtain four results through Beam search, based on which we conduct multi-dimensional evaluation and selection.
    \end{tablenotes}
\label{tab:result of ensemble}
\end{table}

\begin{table*}[htbp]
\caption{Comparison of models on the MEDIQA and MeqSum datasets.}
\centering
{\fontsize{3pt}{4pt}\selectfont
\resizebox{\linewidth}{!}{%
\begin{tabular}{l|c|c|c|c|c|c}
\hline
\multirow{2}{*}{\textbf{Model}} & \multicolumn{3}{c|}{\textbf{MEDIQA}} & \multicolumn{3}{c}{\textbf{MeqSum}} \\
\cline{2-7}
 & \textbf{ROUGE-L} & \textbf{BERTscore} & \textbf{$\textsc{SummaC}_\text{ZS}$} & \textbf{ROUGE-L} & \textbf{BERTscore} & \textbf{$\textsc{SummaC}_\text{ZS}$} \\
\hline
BART \cite{lewis2019bart} &0.293&0.747&0.448&0.502&0.817&0.392\\
ECL\cite{wei2023medical} &0.280&0.741&0.477&0.513&0.823&0.393\\
MvCL\cite{wei2024multi}&0.278&0.739&0.476&0.514&0.829&0.407\\
QFCL  \cite{zhang2022focus} &0.289&0.746&0.488&0.522&0.833&0.427\\
Damo\textsuperscript{*}  \cite{he2021damo_nlp} &0.313& - &- &- &- &-  \\
MEDAL\textsuperscript{*} \cite{li2024better} &0.302 &0.754 &0.496 &- &- &-  \\
FaMeSumm\textsuperscript{*}  \cite{zhang2023famesumm} &0.301 &0.753 &0.472 &- &- &- \\
\hline
FocusMed(w/o selection) & \underline{0.364}   & \underline{0.775}& \underline{0.577} & \underline{0.577} & \underline{0.856} & \underline{0.449} \\
FocusMed&\textbf{0.386}&\textbf{0.782}&\textbf{0.603}&\textbf{0.589}&\textbf{0.859}&\textbf{0.473} \\
\hline
\end{tabular}%
}}
\begin{tablenotes}
    	\item       \textbf{Notes}: Due to the lack of a fixed split in MeqSum, we re-ran some baseline models. The remaining results are quoted from the original paper (results marked with *). ``FocusMed (no selection)'' refers to the best-performing result among the four models used in the selection. 
    \end{tablenotes}
\label{tab:comparison_consistent}
\end{table*} 

\subsubsection{Comparison with SOTA Methods}
To demonstrate the effectiveness of our proposed framework, we select several state-of-the-art methods as baselines for comparison. ECL \cite{wei2023medical} and QFCL \cite{zhang2022focus} utilize medical entities and question focus, respectively, to generate hard negative samples and enhance summary quality through contrastive learning. MvCL \cite{wei2024multi} adopts a multi-view similarity reranking strategy to alleviate exposure bias between training and inference stages. Damo \cite{he2021damo_nlp} integrates pre-trained generation, knowledge-base error correction, and output reranking to achieve significant performance gains. MEDAL \cite{li2024better} addresses summary hallucination via a model-agnostic post-processing mechanism. FaMeSumm \cite{zhang2023famesumm} incorporates contrastive learning and medical knowledge to fine-tune pre-trained language models, thereby improving the factual consistency and reliability of the generated summaries.

Table \ref{tab:comparison_consistent} presents the experimental results of our complete framework. It is evident that across both datasets, our framework attains the highest performance across all evaluation metrics, demonstrating its superior effectiveness.

On the MEDIQA dataset, the MEDAL method employs a model-agnostic post-processing framework to detect and correct hallucinated content in clinical summaries through rule matching, semantic analysis, and retrieval-augmented generation, thereby enhancing the faithfulness of the model's outputs. Although this approach significantly improves model performance, it requires the additional training of a model to generate training data. In contrast, our framework not only surpasses the state-of-the-art in terms of performance but also can be easily applied to all datasets related to MQS. In particular, the improvement of the $\textsc{SummaC}_\text{ZS}$ indicator can prove that the effectiveness of our framework in improving the faithfulness of the results far exceeds that of MEDAL.

On the MeqSum dataset, QFCL's results have always been at the leading level. It uses question focus to drive the generation of difficult negative samples and uses contrastive learning to improve sentence-level representation capabilities, thereby generating higher-quality medical quesiton summaries. Compared to this model, our method has achieved better results in all indicators, improving by 6.7\% in ROUGE-L, 2.6\% in BERTscore and 4.6\% in $\textsc{SummaC}_\text{ZS}$.

\section{Discussion}
\begin{figure}[b]
\centering
\includegraphics[width=0.5\textwidth]{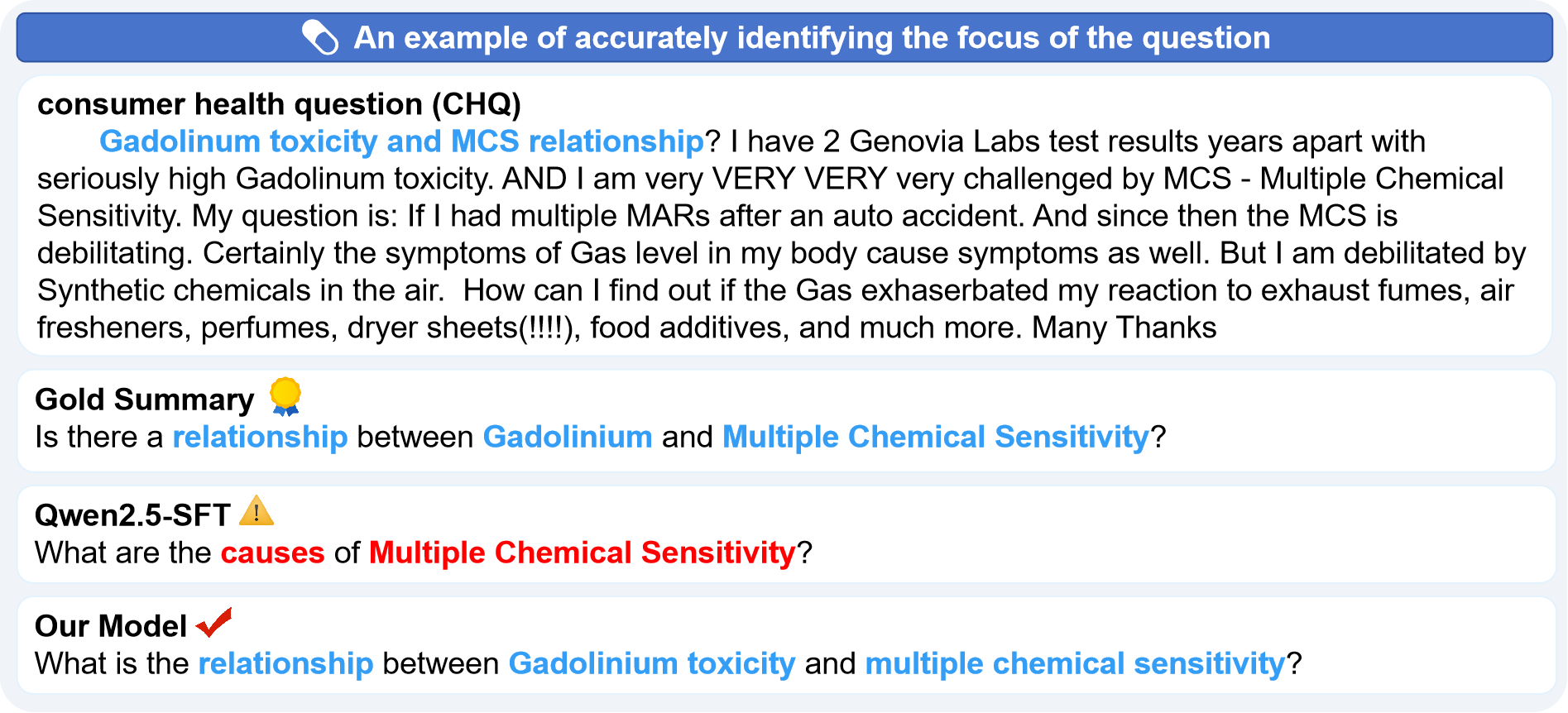}
\caption{An example of accurately identifying the focus of the question. Blue font represents the intended core focus. Red font indicates cases where the model either incorrectly identifies the question focus or fails to capture it completely.}
\label{fig:success case}
\end{figure}
Although the results on two open-source datasets demonstrate the effectiveness of our proposed method, certain limitations still remain in specific scenarios. To better understand the strengths and weaknesses of the FocusMed framework, we further conduct a case study analysis, including one successful example and one failure case.

Accurately identifying the key focus in the user's question is a significant challenge in the MQS task. This not only helps the model better understand the user's intent but also provides a clear starting point for the doctor's response. As illustrated in Fig. \ref{fig:success case}, the patient's concern revolves around \begin{figure}[t]
\centering
\includegraphics[width=0.5\textwidth]{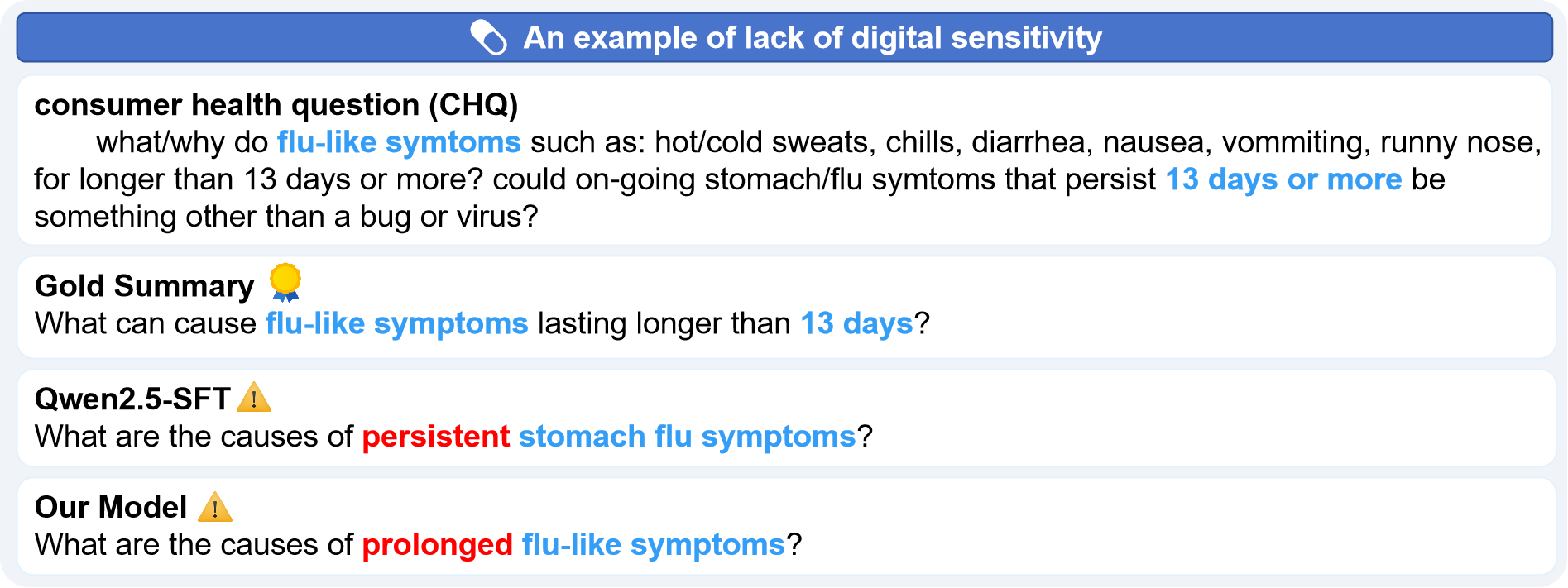}
\caption{An example of a mistake stemming from excessive attention to detail. Blue font represents the intended core focus. Red font indicates cases where the model fails to provide a specific time span.}
\label{fig:failure case}
\end{figure}the relationship between ``Gadolinium toxicity'' and ``Multiple Chemical Sensitivity''. However, the fine-tuned Qwen model mistakenly summarized the patient's intent as exploring the causes of ``Multiple Chemical Sensitivity'', which significantly deviates from the original user intent and hinders physicians from providing satisfactory answers. In contrast, our model's response is concise and clear, explicitly addressing the connection between ``Gadolinium toxicity'' and ``Multiple Chemical Sensitivity'', and seeking to understand their potential association. This greatly facilitates the physician's diagnostic process.

In medical texts, numbers often contain critical information such as patient age, medication dosage, or disease duration, all of which directly impact the accuracy and effectiveness of medical diagnosis. However, due to the relatively low sensitivity of LLMs to numerical details, these models often struggle to focus on and accurately capture key numerical information within the text. For example, in the gold standard summary ``What can cause flu-like symptoms lasting longer than 13 days?'', the crucial number ``13 days'' provides essential context, serving as a vital reference for physicians to assess symptom duration. Although both the Qwen-SFT and our models use terms like ``prolonged'' or ``persistent'' to indicate extended duration, the lack of specific numerical reference may still hinder physicians’ precise evaluation of symptom length, thereby affecting the diagnostic process and potentially impeding the provision of accurate, targeted treatment plans that meet patient expectations.

\section{Conclusion}
This paper systematically investigates the performance of LLMs on the MQS task and proposes an effective optimization strategy that leverages their extensive pre-training corpora to enhance task-specific performance. The approach aims to address common issues such as insufficient question focus extraction and hallucination phenomena. We explore the semantic potential of pre-trained corpora and build an augmented dataset to facilitate model calibration and alignment. Through a faithfulness verification mechanism, we effectively mitigate the hallucination quesiton commonly observed in LLMs. Additionally, we design a multi-dimensional quality evaluation and selection mechanism tailored for summarization tasks, which further enhances the overall system performance. Experimental results show that the proposed system achieves significant improvements on two MQS datasets and sets a new state-of-the-art. However, despite the generally positive performance of our framework, there are still some limitations, such as overemphasis on details and insufficient sensitivity to critical data like numbers and time in certain scenarios. These issues may affect the accuracy of the final results. In the future, we will focus on further optimizing the model's handling of details and enhancing its sensitivity to key information.

\section*{Acknowledgment}
This research was supported by the Natural Science Foundation of China (No. 62302076, 62276043), the Fundamental Research Funds for the Central Universities (No. DUT25YG108), and the Research Project on High Quality Development of Hospital Pharmacy, National Institute of Hospital Administration, NHC, China (No. NIHAYSZX2525).
\bibliographystyle{IEEEtran}

\begin{thebibliography}{10}
\providecommand{\url}[1]{#1}
\csname url@samestyle\endcsname
\providecommand{\newblock}{\relax}
\providecommand{\bibinfo}[2]{#2}
\providecommand{\BIBentrySTDinterwordspacing}{\spaceskip=0pt\relax}
\providecommand{\BIBentryALTinterwordstretchfactor}{4}
\providecommand{\BIBentryALTinterwordspacing}{\spaceskip=\fontdimen2\font plus
\BIBentryALTinterwordstretchfactor\fontdimen3\font minus \fontdimen4\font\relax}
\providecommand{\BIBforeignlanguage}[2]{{%
\expandafter\ifx\csname l@#1\endcsname\relax
\typeout{** WARNING: IEEEtran.bst: No hyphenation pattern has been}%
\typeout{** loaded for the language `#1'. Using the pattern for}%
\typeout{** the default language instead.}%
\else
\language=\csname l@#1\endcsname
\fi
#2}}
\providecommand{\BIBdecl}{\relax}
\BIBdecl

\bibitem{abacha2019summarization}
A.~B. Abacha and D.~Demner-Fushman, ``On the summarization of consumer health questions,'' in \emph{Proceedings of the 57th Annual Meeting of the Association for Computational Linguistics}, 2019, pp. 2228--2234.

\bibitem{sutskever2014sequence}
I.~Sutskever, O.~Vinyals, and Q.~V. Le, ``Sequence to sequence learning with neural networks,'' \emph{Advances in neural information processing systems}, vol.~27, 2014.

\bibitem{liu2019text}
Y.~Liu and M.~Lapata, ``Text summarization with pretrained encoders,'' \emph{arXiv preprint arXiv:1908.08345}, 2019.

\bibitem{lewis2019bart}
M.~Lewis, Y.~Liu, N.~Goyal, M.~Ghazvininejad, A.~Mohamed, O.~Levy, V.~Stoyanov, and L.~Zettlemoyer, ``Bart: Denoising sequence-to-sequence pre-training for natural language generation, translation, and comprehension,'' \emph{arXiv preprint arXiv:1910.13461}, 2019.

\bibitem{zhang2022focus}
M.~Zhang, S.~Dou, Z.~Wang, and Y.~Wu, ``Focus-driven contrastive learniang for medical question summarization,'' \emph{arXiv preprint arXiv:2209.00484}, 2022.

\bibitem{yadav2021reinforcement}
S.~Yadav, D.~Gupta, A.~B. Abacha, and D.~Demner-Fushman, ``Reinforcement learning for abstractive question summarization with question-aware semantic rewards,'' \emph{arXiv preprint arXiv:2107.00176}, 2021.

\bibitem{wei2023medical}
S.~Wei, W.~Lu, X.~Peng, S.~Wang, Y.-F. Wang, and W.~Zhang, ``Medical question summarization with entity-driven contrastive learning,'' \emph{arXiv preprint arXiv:2304.07437}, 2023.

\bibitem{achiam2023gpt}
J.~Achiam, S.~Adler, S.~Agarwal, L.~Ahmad, I.~Akkaya, F.~L. Aleman, D.~Almeida, J.~Altenschmidt, S.~Altman, S.~Anadkat \emph{et~al.}, ``Gpt-4 technical report,'' \emph{arXiv preprint arXiv:2303.08774}, 2023.

\bibitem{singhal2025toward}
K.~Singhal, T.~Tu, J.~Gottweis, R.~Sayres, E.~Wulczyn, M.~Amin, L.~Hou, K.~Clark, S.~R. Pfohl, H.~Cole-Lewis \emph{et~al.}, ``Toward expert-level medical question answering with large language models,'' \emph{Nature Medicine}, vol.~31, no.~3, pp. 943--950, 2025.

\bibitem{luo2024taiyi}
L.~Luo, J.~Ning, Y.~Zhao, Z.~Wang, Z.~Ding, P.~Chen, W.~Fu, Q.~Han, G.~Xu, Y.~Qiu \emph{et~al.}, ``Taiyi: a bilingual fine-tuned large language model for diverse biomedical tasks,'' \emph{Journal of the American Medical Informatics Association}, vol.~31, no.~9, pp. 1865--1874, 2024.

\bibitem{yu2024enhancing}
H.~Yu, C.~Yu, Z.~Wang, D.~Zou, and H.~Qin, ``Enhancing healthcare through large language models: A study on medical question answering,'' in \emph{2024 IEEE 6th International Conference on Power, Intelligent Computing and Systems (ICPICS)}.\hskip 1em plus 0.5em minus 0.4em\relax IEEE, 2024, pp. 895--900.

\bibitem{diekmann2025evaluating}
Y.~Diekmann, C.~M. Fensore, R.~M. Carrillo-Larco, N.~Pradhan, B.~Appana, and J.~C. Ho, ``Evaluating safety of large language models for patient-facing medical question answering,'' in \emph{Proceedings of the 4th Machine Learning for Health Symposium}, vol. 259, 2025, pp. 267--290.

\bibitem{nori2023capabilities}
H.~Nori, N.~King, S.~M. McKinney, D.~Carignan, and E.~Horvitz, ``Capabilities of gpt-4 on medical challenge problems,'' \emph{arXiv preprint arXiv:2303.13375}, 2023.

\bibitem{schumacher2023extrinsically}
E.~Schumacher, D.~Rosenthal, V.~Nair, L.~Price, G.~Tso, and A.~Kannan, ``Extrinsically-focused evaluation of omissions in medical summarization,'' \emph{arXiv preprint arXiv:2311.08303}, 2023.

\bibitem{yang2025qwen3}
A.~Yang, A.~Li, B.~Yang, B.~Zhang, B.~Hui, B.~Zheng, B.~Yu, C.~Gao, C.~Huang, C.~Lv \emph{et~al.}, ``Qwen3 technical report,'' \emph{arXiv preprint arXiv:2505.09388}, 2025.

\bibitem{pham2024towards}
D.~K. Pham and B.~Q. Vo, ``Towards reliable medical question answering: Techniques and challenges in mitigating hallucinations in language models,'' \emph{arXiv preprint arXiv:2408.13808}, 2024.

\bibitem{pal2023med}
A.~Pal, L.~K. Umapathi, and M.~Sankarasubbu, ``Med-halt: Medical domain hallucination test for large language models,'' \emph{arXiv preprint arXiv:2307.15343}, 2023.

\bibitem{zuo2024medhallbench}
K.~Zuo and Y.~Jiang, ``Medhallbench: A new benchmark for assessing hallucination in medical large language models,'' \emph{arXiv preprint arXiv:2412.18947}, 2024.

\bibitem{dettmers2023qlora}
T.~Dettmers, A.~Pagnoni, A.~Holtzman, and L.~Zettlemoyer, ``Qlora: Efficient finetuning of quantized llms,'' \emph{Advances in neural information processing systems}, vol.~36, pp. 10\,088--10\,115, 2023.

\bibitem{savage2024fine}
T.~Savage, S.~Ma, A.~Boukil, V.~Patel, E.~Rangan, I.~Lopez, and J.~H. Chen, ``Fine tuning large language models for medicine: The role and importance of direct preference optimization,'' \emph{arXiv preprint arXiv:2409.12741}, 2024.

\bibitem{xie2023doclens}
Y.~Xie, S.~Zhang, H.~Cheng, P.~Liu, Z.~Gero, C.~Wong, T.~Naumann, H.~Poon, and C.~Rose, ``Doclens: Multi-aspect fine-grained evaluation for medical text generation,'' \emph{arXiv preprint arXiv:2311.09581}, 2023.

\bibitem{mihalcea2004textrank}
R.~Mihalcea and P.~Tarau, ``Textrank: Bringing order into text,'' in \emph{Proceedings of the 2004 conference on empirical methods in natural language processing}, 2004, pp. 404--411.

\bibitem{dubey2024llama}
A.~Dubey, A.~Jauhri, A.~Pandey, A.~Kadian, A.~Al-Dahle, A.~Letman, A.~Mathur, A.~Schelten, A.~Yang, A.~Fan \emph{et~al.}, ``The llama 3 herd of models,'' \emph{arXiv e-prints}, pp. arXiv--2407, 2024.

\bibitem{guo2025deepseek}
D.~Guo, D.~Yang, H.~Zhang, J.~Song, R.~Zhang, R.~Xu, Q.~Zhu, S.~Ma, P.~Wang, X.~Bi \emph{et~al.}, ``Deepseek-r1: Incentivizing reasoning capability in llms via reinforcement learning,'' \emph{arXiv preprint arXiv:2501.12948}, 2025.

\bibitem{abacha2021overview}
A.~B. Abacha, Y.~M’rabet, Y.~Zhang, C.~Shivade, C.~Langlotz, and D.~Demner-Fushman, ``Overview of the mediqa 2021 shared task on summarization in the medical domain,'' in \emph{Proceedings of the 20th workshop on biomedical language processing}, 2021, pp. 74--85.

\bibitem{lin2004rouge}
C.-Y. Lin, ``Rouge: A package for automatic evaluation of summaries,'' in \emph{Text summarization branches out}, 2004, pp. 74--81.

\bibitem{zhang2019bertscore}
T.~Zhang, V.~Kishore, F.~Wu, K.~Q. Weinberger, and Y.~Artzi, ``Bertscore: Evaluating text generation with bert,'' \emph{arXiv preprint arXiv:1904.09675}, 2019.

\bibitem{laban2022summac}
P.~Laban, T.~Schnabel, P.~N. Bennett, and M.~A. Hearst, ``Summac: Re-visiting nli-based models for inconsistency detection in summarization,'' \emph{Transactions of the Association for Computational Linguistics}, vol.~10, pp. 163--177, 2022.

\bibitem{wei2024multi}
S.~Wei, X.~Peng, H.~Guan, L.~Geng, P.~Jian, H.~Wu, and W.~Lu, ``Multi-view contrastive learning for medical question summarization,'' in \emph{2024 27th International Conference on Computer Supported Cooperative Work in Design (CSCWD)}.\hskip 1em plus 0.5em minus 0.4em\relax IEEE, 2024, pp. 1826--1831.

\bibitem{he2021damo_nlp}
Y.~He, M.~Chen, and S.~Huang, ``damo\_nlp at mediqa 2021: knowledge-based preprocessing and coverage-oriented reranking for medical question summarization,'' in \emph{Proceedings of the 20th Workshop on Biomedical Language Processing}, 2021, pp. 112--118.

\bibitem{li2024better}
S.~Li, Y.~Zhang, C.~Deng, Y.~Niu, and H.~Zhao, ``Better late than never: Model-agnostic hallucination post-processing framework towards clinical text summarization,'' in \emph{Findings of the Association for Computational Linguistics ACL 2024}, 2024, pp. 995--1011.

\bibitem{zhang2023famesumm}
N.~Zhang, Y.~Zhang, W.~Guo, P.~Mitra, and R.~Zhang, ``Famesumm: Investigating and improving faithfulness of medical summarization,'' \emph{arXiv preprint arXiv:2311.02271}, 2023.

\end{thebibliography}

\end{document}